\documentclass[10pt, a4paper]{article}
\usepackage{lrec}
\usepackage{graphicx}
\usepackage{tabularx}
\usepackage{soul}
\usepackage{color}
\usepackage[latin1]{inputenc}
\usepackage{hyperref}
\usepackage{xstring}
\usepackage{amsmath}
\usepackage{pifont}
\usepackage{amssymb}
\usepackage{caption}
\usepackage{subfig}

\newcommand{\qu}[1]{``{#1}''}
\newcommand{\wasabinavigator}{WASABI Interactive Navigator}
\newcommand{\wasabicorpus}{WASABI Song Corpus}
\newcommand{\wasabi}{WASABI Song Corpus}
\newcommand{\wasabigithub}{https://github.com/micbuffa/WasabiDataset}

\title{Love Me, Love Me, Say (and Write!) that You Love Me:\\ Enriching the WASABI Song Corpus with Lyrics Annotations}
\name{Michael Fell, Elena Cabrio, Elmahdi Korfed, Michel Buffa and Fabien Gandon}

\address{Universit\'e C\^ote d'Azur, CNRS, Inria, I3S, France \\
          \{michael.fell, elena.cabrio, elmahdi.korfed, michel.buffa\}@unice.fr,  fabien.gandon@inria.fr\\
        }

\abstract{
We present the \wasabicorpus{}, a large corpus of songs enriched with metadata extracted from  music databases on  the  Web, and resulting from the processing of song lyrics and from audio analysis. More specifically, given that lyrics encode an important part of the semantics of a song, we focus here on the description of the methods we proposed to extract relevant information from the lyrics, such as their structure segmentation, their topics, the explicitness of the lyrics content, the salient passages of a song and the emotions conveyed. The creation of the resource is still ongoing: so far, the corpus contains 1.73M songs with lyrics (1.41M unique lyrics)  annotated at different levels with the output of the above mentioned methods. Such corpus labels and the provided methods can be exploited by music search engines and music professionals (e.g. journalists, radio presenters) to better handle large collections of lyrics, allowing an intelligent browsing, categorization and recommendation of songs. We provide the files of the current version of the \wasabicorpus{}, the models we have built on it as well as updates here: \url{\wasabigithub}.
\\ \newline  }

\begin{document}

 \maketitleabstract

\section{Introduction}
\label{sec:introduction}
Let's imagine the following scenario: following David Bowie's death, a journalist plans to prepare a radio show about the artist's musical career to acknowledge his qualities. To discuss the topic from different angles, she needs to have at her disposal the artist biographical information to know the history of his career, the song lyrics to know what he was singing about, his musical style, the emotions his songs were conveying, live recordings and interviews.
Similarly, streaming professionals such as Deezer, Spotify, Pandora or Apple Music aim at enriching music listening with artists' information, to offer suggestions for listening to other songs/albums from the same or similar artists, or automatically determining the emotion felt when listening  to  a  track to propose coherent playlists to the user.
To support such scenarios, the need for rich and accurate musical knowledge bases and tools to explore and exploit this knowledge becomes evident.

In this paper, we present the \wasabicorpus{}, a large corpus of songs (2.10M songs, 1.73M with lyrics) enriched with metadata extracted from  music databases on the  Web, and resulting from the processing of song lyrics and from audio analysis. The corpus contains songs in 36 different languages,  even if the  vast  majority are in English.  As for the songs genres, the most common ones are Rock, Pop, Country and Hip Hop.

More specifically, while an overview of the goals of the WASABI  project supporting the dataset creation and the description of a preliminary version of the dataset can be found in \cite{wasabi}, this paper focuses on the description of the methods we proposed to annotate relevant information in the song lyrics. Given that lyrics encode an important part of the semantics of a song, we propose to label the WASABI dataset lyrics with their structure segmentation, the explicitness of the lyrics content, the salient passages of a song, the addressed topics and the emotions conveyed. 

An analysis of the correlations among the above mentioned annotation layers reveals interesting insights about the song corpus. For instance, we demonstrate the change in corpus annotations diachronically: we show that certain topics become more important over time and others are diminished. We also analyze such changes in explicit lyrics content and expressed emotion.

The paper is organized as follows. Section \ref{sec:wasabi_corpus} introduces the \wasabicorpus{} and the metadata initially extracted from music databases on the Web. Section \ref{sec:structure} describes the segmentation method we applied to decompose lyrics in their building blocks in the corpus. Section \ref{sec:summarization} explains the method used to summarize song lyrics, leveraging their structural properties. Section \ref{sec:explictness} reports on the annotations resulting from the explicit content classifier, while Section \ref{sec:emotion} describes how information on the emotions are extracted from the lyrics. Section \ref{sec:topics} describes the topic modeling algorithm to label each lyrics with the top 10 words, while Section \ref{sec:time_analysis} examines the changes in the annotations over the course
of time.
Section \ref{sec:related} reports on similar existing resources, while Section \ref{sec:conclusion} concludes the paper.
\section{The \wasabicorpus{}} \label{sec:wasabi_corpus}
In the context of the WASABI research project\footnote{\url{http://wasabihome.i3s.unice.fr/}} that started in 2017, a two million song database has been built, with metadata on 77k artists, 208k albums, and 2.10M songs \cite{wasabi}. The metadata has been \textit{i)} aggregated, merged and curated from different data sources on the Web, and \textit{ii)} enriched by pre-computed or on-demand analyses of the lyrics and audio data.

\begin{figure}
    \centering
    \includegraphics[width=\linewidth]{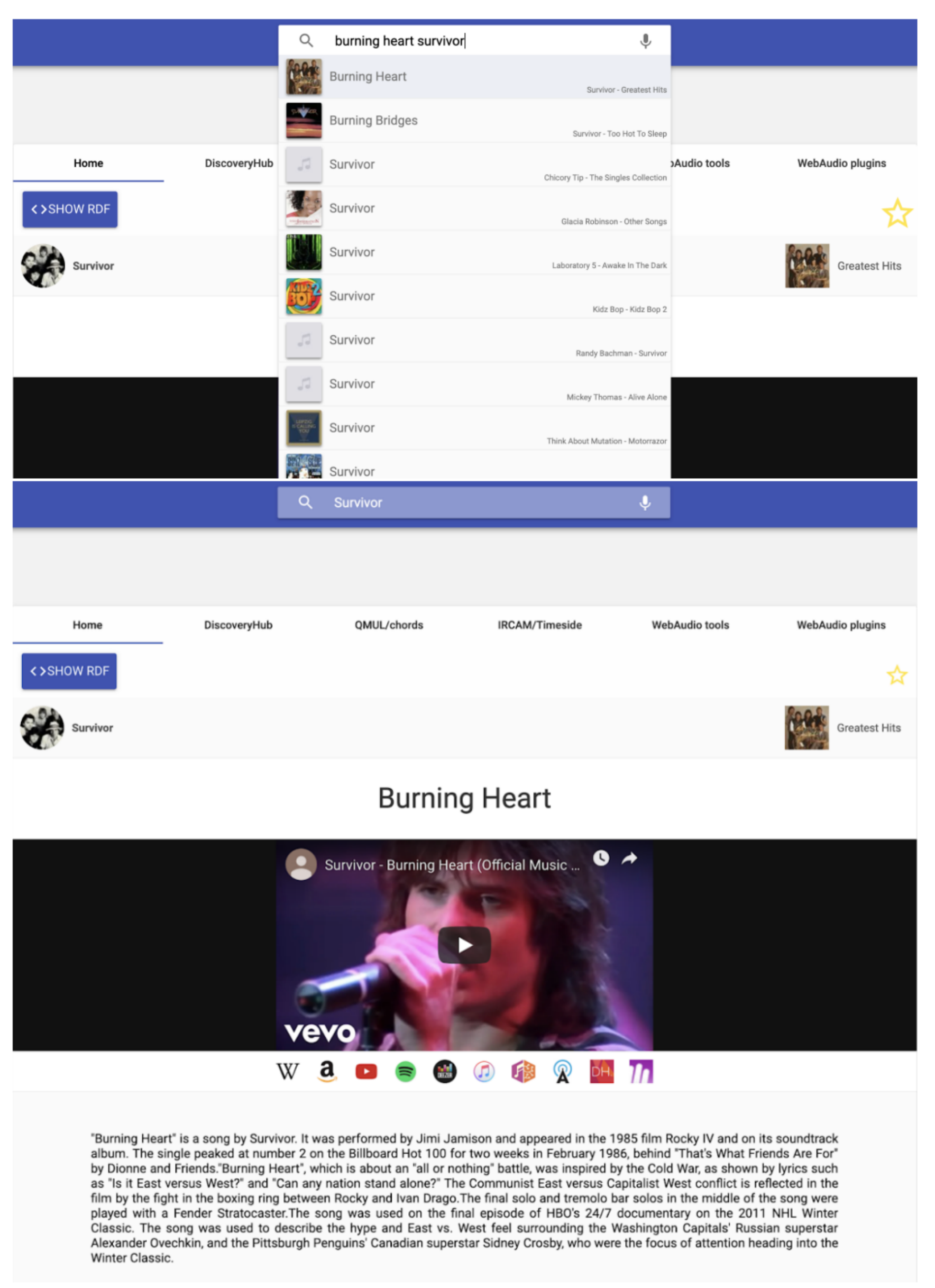}
    \caption{The \wasabinavigator{}.}
    \label{fig:wasabi_project:wasabi_interactive_navigator}
\end{figure}

We have performed various levels of analysis, and interactive Web Audio applications have been built on top of the output. For example, the TimeSide analysis and annotation framework have been linked \cite{fillon2014telemeta} to make on-demand audio analysis possible. In connection with the FAST project\footnote{\url{http://www.semanticaudio.ac.uk}}, an offline chord analysis of 442k songs has been performed, and both an online enhanced audio player \cite{pauwels2019iui} and chord search engine \cite{pauwels2018wac} have been built around it.
A rich set of Web Audio applications and plugins has been proposed \cite{buffa2017wac1,buffa2017wac2,buffa2018towards}, that allow, for example, songs to be played along with sounds similar to those used by artists. All these metadata, computational analyses and Web Audio applications have now been gathered in one easy-to-use web interface, the \wasabinavigator{}\footnote{\url{http://wasabi.i3s.unice.fr/}}, illustrated\footnote{Illustration taken from \cite{wasabi_interactive_navigator}.} in Figure~\ref{fig:wasabi_project:wasabi_interactive_navigator}.

We have started building the \wasabicorpus{} by collecting for each artist the complete discography, band members with their instruments, time line, equipment they use, and so on. For each song we collected its lyrics from LyricWiki\footnote{ \url{http://lyrics.wikia.com/}}, the synchronized lyrics when available\footnote{from \url{http://usdb.animux.de/}}, the DBpedia abstracts and the categories the song belongs to, e.g. genre, label, writer, release date, awards, producers, artist and band members, the stereo audio track from Deezer, the unmixed audio tracks of the song, its ISRC, bpm and duration.

\begin{figure}
    \centering
    \includegraphics[width=\linewidth]{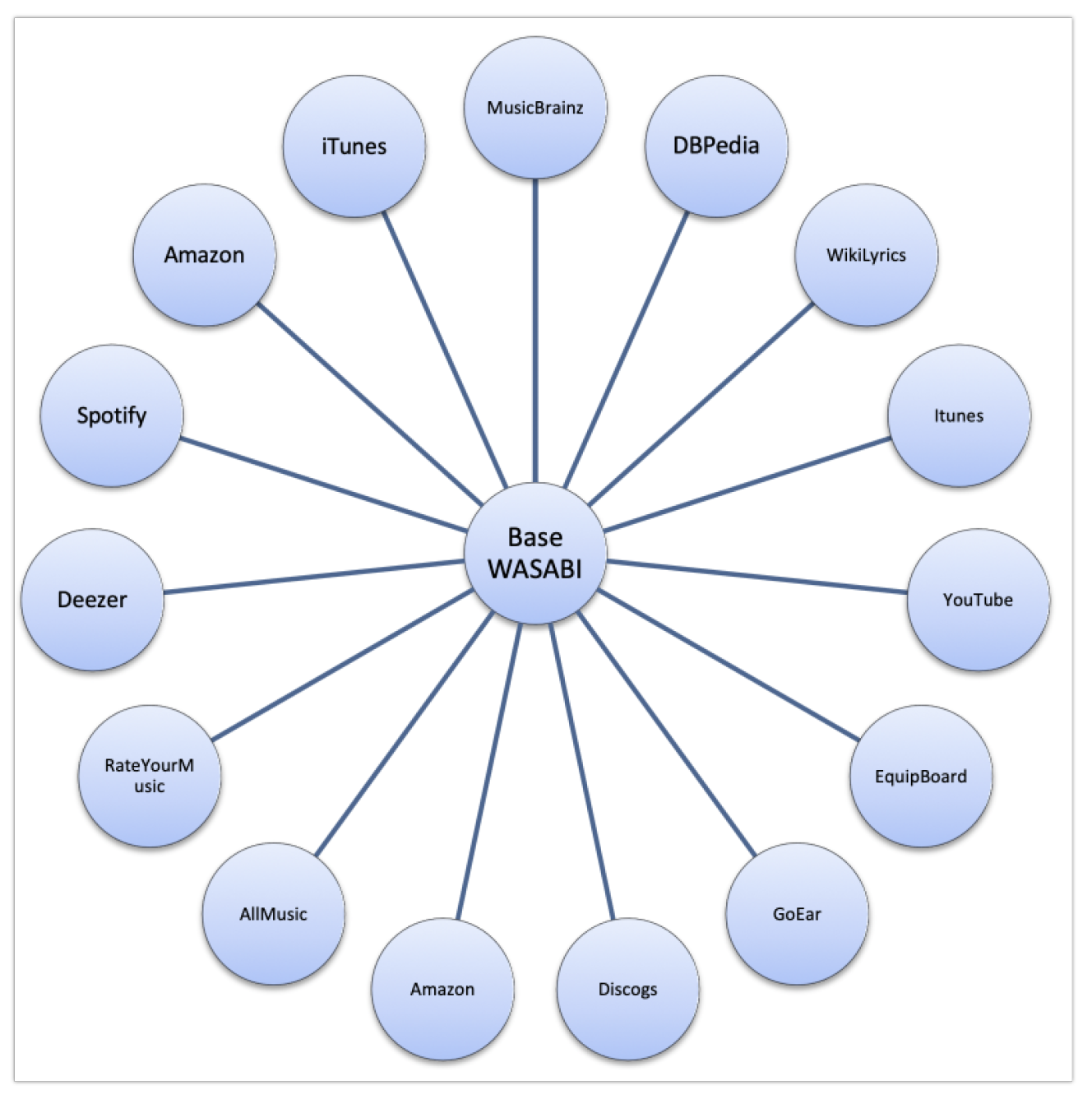}
    \caption{The datasources connected to the \wasabi{}.}
    \label{fig:wasabi_project:wasabi_datasources}
\end{figure}

We matched the song ids from the \wasabicorpus{} with the ids from MusicBrainz, iTunes, Discogs, Spotify, Amazon, AllMusic, GoHear, YouTube. Figure~\ref{fig:wasabi_project:wasabi_datasources} illustrates\footnote{Illustration taken from \cite{buffa2019webaudio}.} all the data sources we have used to create the \wasabicorpus{}. We have also aligned the \wasabicorpus{} with the publicly available LastFM dataset\footnote{\url{http://millionsongdataset.com/lastfm/}}, resulting in 327k tracks in our corpus having a LastFM id.

As of today, the corpus contains 1.73M songs with lyrics (1.41M unique lyrics). 73k songs have at least an abstract on DBpedia, and 11k  have been
identified as \qu{classic songs} (they have been number one, or
got a Grammy award, or have lots of cover versions). About 2k songs have a multi-track audio version, and on-demand source separation using open-unmix \cite{stoter2019open} or Spleeter \cite{spleeter2019} is provided as a TimeSide plugin.

Several Natural Language Processing methods have been applied to the lyrics of the songs included in the \wasabicorpus{}, as well as various analyses of the extracted information have been carried out.
After providing some statistics on the WASABI corpus, the rest of the article describes the different annotations we added to the lyrics of the songs in the dataset. Based on the research we have conducted, the following lyrics annotations are added: lyrical structure (Section~\ref{sec:structure}), summarization (Section~\ref{sec:summarization}), explicit lyrics (Section~\ref{sec:explictness}), emotion in lyrics (Section~\ref{sec:emotion}) and topics in lyrics (Section~\ref{sec:topics}).

\subsection{Statistics on the \wasabicorpus{}}
\begin{figure}
\centering
    \subfloat[Language distribution (100\% = 1.73M)
\label{fig:languages}]{%
      \includegraphics[width=\linewidth]{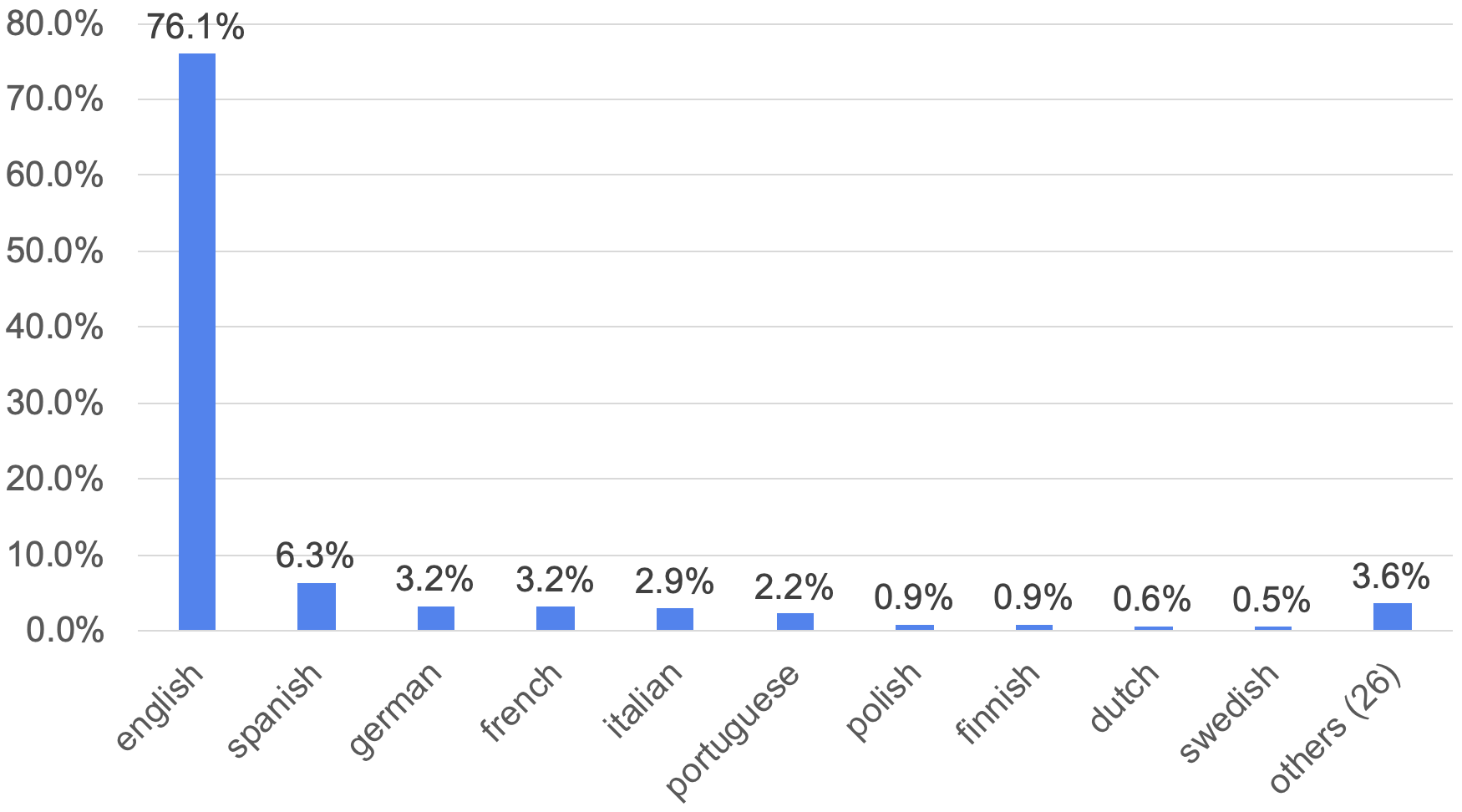}%
      }\par\vspace{4ex}
    \subfloat[Genre distribution (100\% = 1.06M) \label{fig:genres}]{%
      \includegraphics[width=\linewidth]{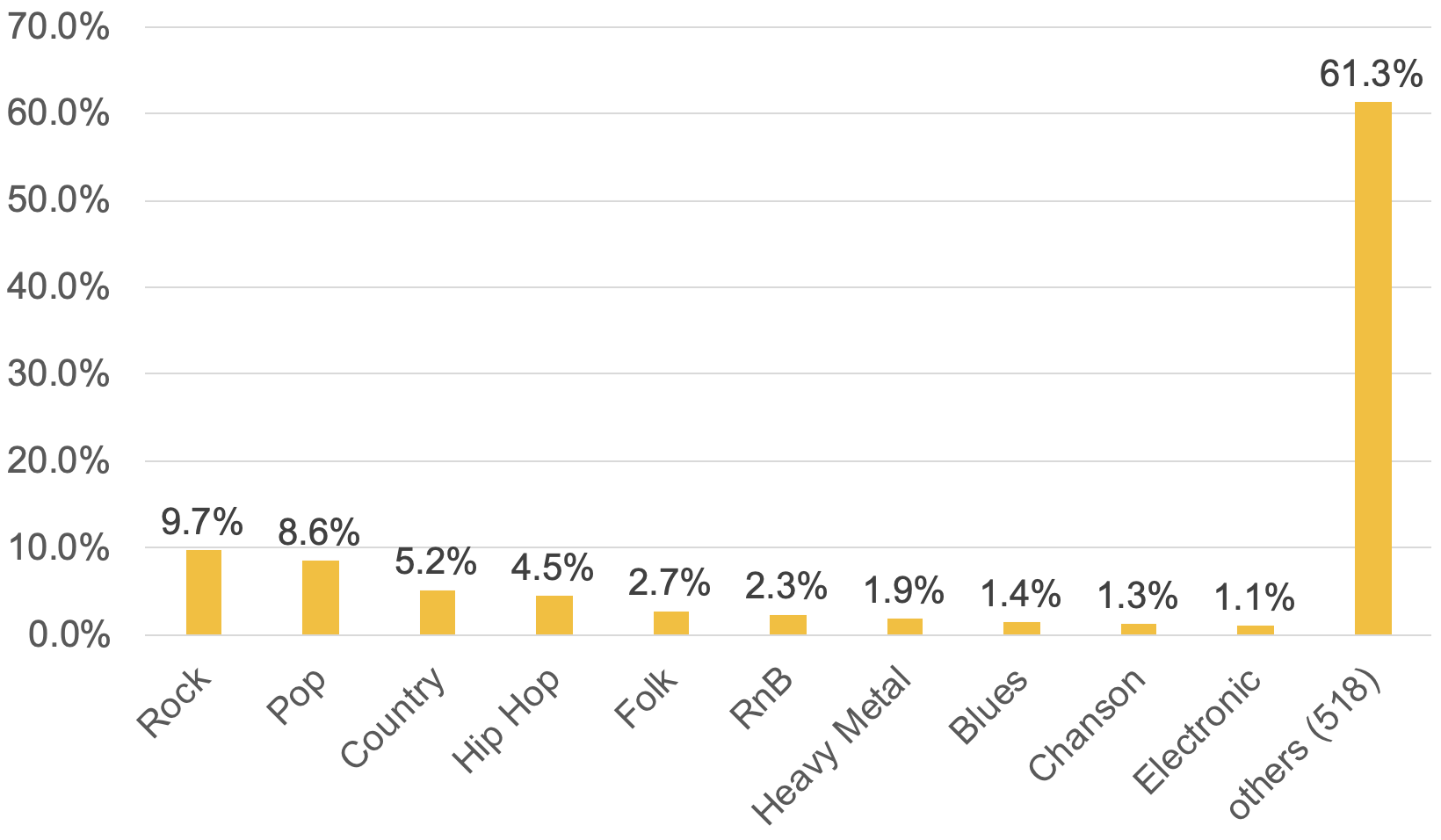}%
      }\par\vspace{4ex}     
    \subfloat[Decade of publication distribution (100\% = 1.70M)
\label{fig:years}]{%
      \includegraphics[width=\linewidth]{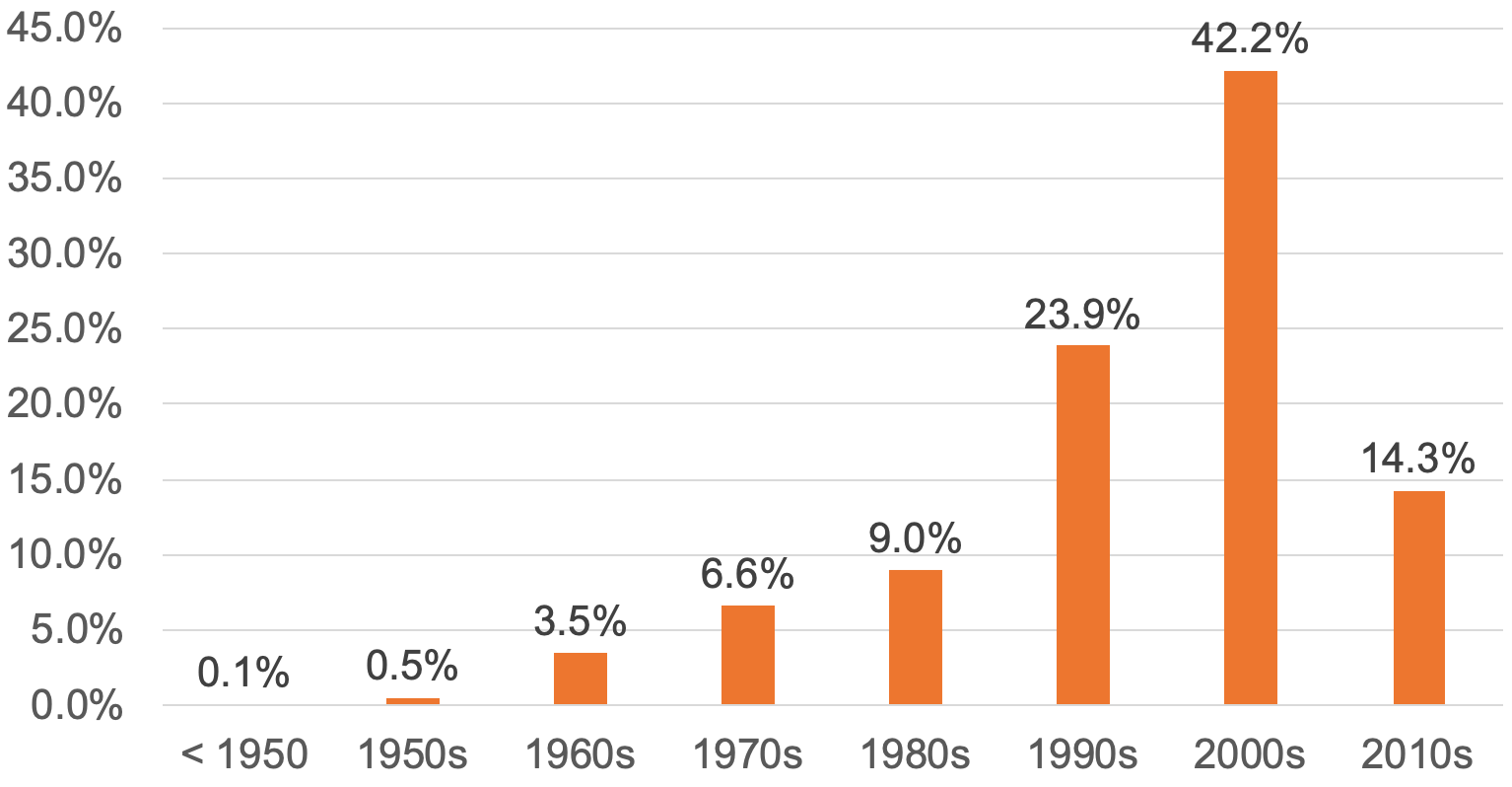}%
      }
\caption{Statistics on the \wasabicorpus{}}
\label{fig:statistics}
\end{figure}

This section summarizes key statistics on the corpus, such as the language and genre distributions, the songs coverage in terms of publication years, and then gives the technical details on its accessibility.
\paragraph{Language Distribution}
Figure~\ref{fig:languages} shows the distribution of the ten most frequent languages in our corpus.\footnote{Based on language detection performed on the lyrics.} In total, the corpus contains songs of 36 different languages. The vast majority (76.1\%) is English, followed by Spanish (6.3\%) and by four languages in the 2-3\% range (German, French, Italian, Portugese). On the bottom end, Swahili and Latin amount to 0.1\% (around 2k songs) each.

\paragraph{Genre Distribution}
In Figure~\ref{fig:genres} we depict the distribution of the ten most frequent genres in the corpus.\footnote{We take the genre of the album as ground truth since song-wise genres are much rarer.} In total, 1.06M of the titles are tagged with a genre. It should be noted that the genres are very sparse with a total of 528 different ones. This high number is partially due to many subgenres such as Alternative Rock, Indie Rock, Pop Rock, etc. which we omitted in Figure~\ref{fig:genres} for clarity. The most common genres are Rock (9.7\%), Pop (8.6\%), Country (5.2\%),  Hip Hop (4.5\%) and Folk (2.7\%).

\paragraph{Publication Year}
Figure~\ref{fig:years} shows the number of songs published in our corpus, by decade.\footnote{We take the album publication date as proxy since song-wise labels are too sparse.} We find that over 50\% of all songs in the \wasabicorpus{} are from the 2000s or later and only around 10\% are from the seventies or earlier.

\paragraph{Accessibility of the \wasabicorpus{}}
The \wasabinavigator{} relies on multiple database engines: it  runs on a MongoDB server altogether with an indexation by Elasticsearch and also on a Virtuoso triple store as a RDF graph database. It comes with a
% rather complete
REST API\footnote{\url{https://wasabi.i3s.unice.fr/apidoc/}} and an upcoming SPARQL endpoint. 
All the database metadata is publicly available\footnote{There is no public
access to copyrighted data such as lyrics and full length
audio files. Instructions on how to obtain lyrics are nevertheless provided and audio extracts of 30s length are
available for nearly all songs.} under a CC licence through the \wasabinavigator{} as well as programmatically
through the WASABI REST API.

We provide the files of the current version of the \wasabicorpus{}, the models we have built on it as well as updates here: \url{\wasabigithub}.
\section{Lyrics Structure Annotations}
\label{sec:structure}
Generally speaking, lyrics structure segmentation consists of two stages: text segmentation to divide lyrics into segments, and semantic labelling to label each segment with a structure type (e.g. Intro, Verse, Chorus).

\begin{figure}
    \centering
    \includegraphics[width=\linewidth]{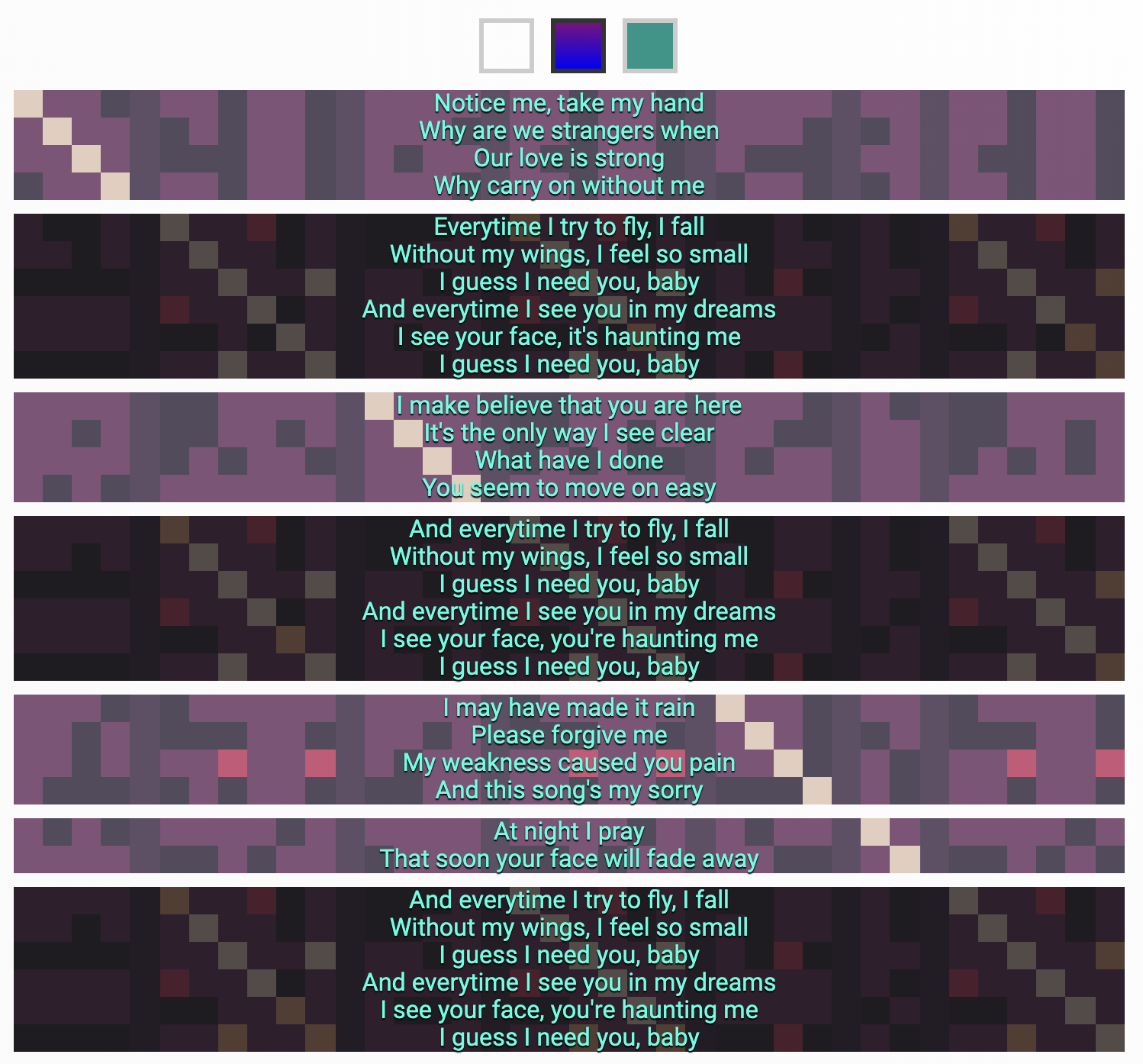}
    \caption{Structure of the lyrics of \qu{Everytime} by Britney Spears as displayed in the \wasabinavigator{}.}
    \label{fig:structure}
\end{figure}

In \cite{fell:segmentation} we proposed a method to segment lyrics based on their repetitive structure in the form of a self-similarity matrix (SSM). Figure~\ref{fig:structure} shows a line-based SSM for the song text written on top of it\footnote{\url{https://wasabi.i3s.unice.fr/\#/search/artist/Britney\%20Spears/album/In\%20The\%20Zone/song/Everytime}}. The lyrics consists of seven segments and shows the typical repetitive structure of a Pop song. The main diagonal is trivial, since each line is maximally similar to itself. Notice further the additional diagonal stripes in segments 2, 4 and 7; this indicates a repeated part, typically the chorus. Based on the simple idea that eyeballing an SSM will reveal (parts of) a song's structure, we proposed a Convolutional Neural Network architecture that successfully learned to predict segment borders in the lyrics when \qu{looking at} their SSM. Table~\ref{tab:segmentation:result_genrewise} shows the genre-wise results we obtained using our proposed architecture. One important insight was that more repetitive lyrics as often found in genres such as Country and Punk Rock are much easier to segment than lyrics in Rap or Hip Hop which often do not even contain a chorus.

\begin{table}
    \centering
    \begin{tabular}{lccc} \hline
\textit{Genre} & \textit{P} & \textit{R} & $F_1$ \\ \hline			
Rock                  & 73.8 & 	57.7 & 	64.8 \\ 
Hip Hop	              & 71.7 &	43.6 &	\underline{54.2} \\ 
Pop	                  & 73.1 &	61.5 &	66.6 \\ 
RnB	                  & 71.8 &	60.3 &	65.6 \\ 
Alternative Rock      & 76.8 &	60.9 &	67.9 \\ 
Country	               & 74.5 &	66.4 &	\textbf{70.2} \\ 
Hard Rock              & 76.2 &	61.4 &	67.7 \\ 
Pop Rock               & 73.3 & 	59.6 &	65.8 \\ 
Indie Rock	           & 80.6 &	55.5 &	65.6 \\ 
Heavy Metal	           & 79.1 &	52.1 &	63.0 \\ 
Southern Hip Hop        & 73.6 &	34.8 &	\underline{47.0} \\ 
Punk Rock               & 80.7 &	63.2 &	\textbf{70.9} \\ 
Alternative Metal       & 77.3 &	61.3 &	68.5 \\ 
Pop Punk                & 77.3 &	68.7 &	\textbf{72.7} \\ 
Gangsta Rap             & 73.6 &	35.2 &	\underline{47.7} \\
Soul                    & 70.9 &	57.0 &	63.0 \\ 
 \hline
    \end{tabular}
    \caption{Lyrics segmentation performances across musical genres in terms of Precision (\textit{P}), Recall (\textit{R}) and $F_1$ in \%. Underlined are the performances on genres with less repetitive text. Genres with highly repetitive structure are in bold.}
    \label{tab:segmentation:result_genrewise}
\end{table}

In the \wasabinavigator{}, the line-based SSM of a song text can be visualized. It is toggled by clicking on the violet-blue square on top of the song text. For a subset of songs the color opacity indicates how repetitive and representative a segment is, based on the fitness metric that we proposed in \cite{fell:summarization}. Note how in Figure~\ref{fig:structure}, the segments 2, 4 and 7 are shaded more darkly than the surrounding ones. As highly fit (opaque) segments often coincide with a chorus, this is a first approximation of chorus detection. Given the variability in the set of structure types provided in the literature according to different genres \cite{tagg_1982,brackett1995interpreting}, rare attempts have been made in the literature to achieve a more complete semantic labelling, labelling the lyrics segments as Intro, Verse, Bridge, Chorus etc.

For each song text we provide an SSM based on a normalized character-based edit distance\footnote{In our segmentation experiments we found this simple metric to outperform more complex metrics that take into account the phonetics or the syntax.} on two levels of granularity to enable other researchers to work with these structural representations: line-wise similarity and segment-wise similarity.
\section{Lyrics Summary}
\label{sec:summarization}

\newcommand{\textrank}{$\mathbb{R}ank$}
\newcommand{\topsum}{$\mathbb{T}opic$}
\newcommand{\textranktopsum}{\textrank\topsum}
\newcommand{\fitness}{$\mathbb{F}it$}
\newcommand{\thumbnail}{\fitness{}}
\newcommand{\combined}{\textrank\topsum\fitness}

\begin{figure}
    \centering
    \includegraphics[width=\linewidth]{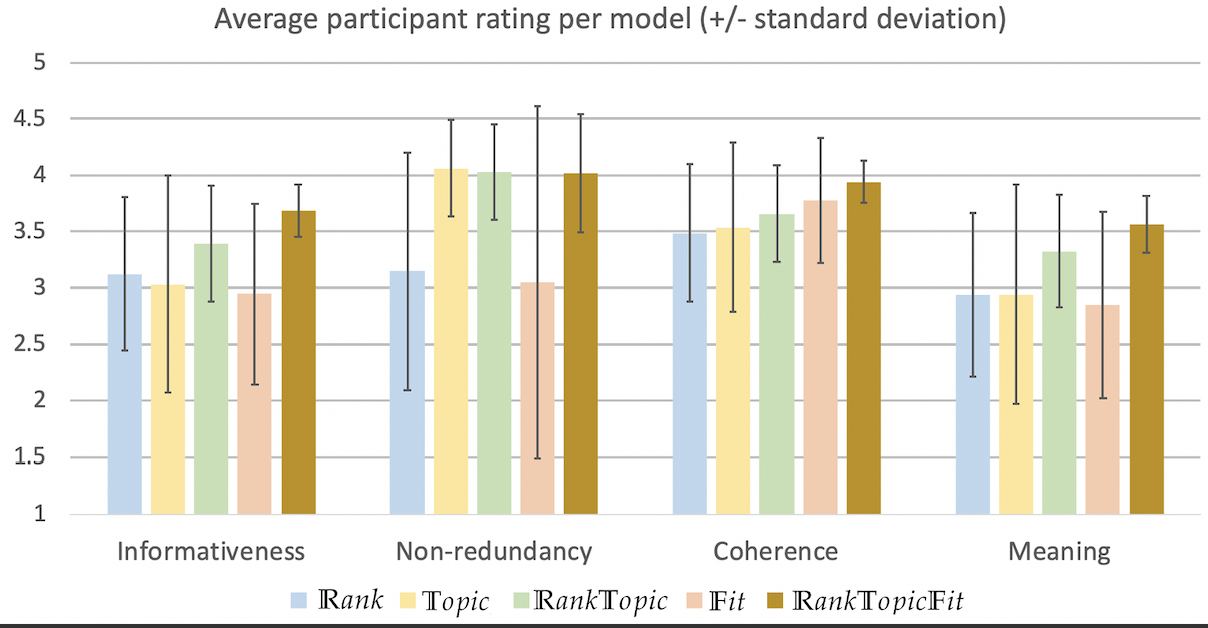}
    \caption{Human ratings per summarization model (five point Likert scale). Models are \textrank{}: graph-based, \topsum{}: topic-based, \thumbnail{}: thumbnail-based, and model combinations.}
    \label{fig:summarization:human_ratings}
\end{figure}

\begin{figure}
    \centering
    \includegraphics[width=6cm]{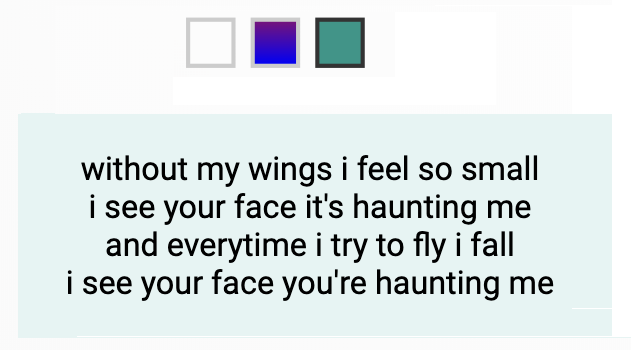}
    \caption{Summary of the lyrics of \qu{Everytime} by Britney Spears as displayed in the \wasabinavigator{}.}
    \label{fig:summary}
\end{figure}

Given the repeating forms, peculiar structure and other unique characteristics of song lyrics, in \cite{fell:summarization} we introduced a method for extractive summarization of lyrics that takes advantage of these additional elements %(i.e. of the context)
to more accurately identify relevant information in song lyrics. More specifically, it relies on the intimate relationship between the audio and the lyrics. The so-called audio thumbnails, snippets of usually 30 seconds of music, are a popular means to summarize a track in the audio community. The intuition is the more repeated and the longer a part, the better it represents the song. We transferred an audio thumbnailing approach to our domain of lyrics and showed that adding the thumbnail improves summary quality. We evaluated our method on 50k lyrics belonging to the top 10 genres of the \wasabicorpus{} and according to qualitative criteria such as \textit{Informativeness} and \textit{Coherence}.
%We selected the lyrics that we estimate to contain a chorus of at least four lines. 
Figure~\ref{fig:summarization:human_ratings} shows our results for different summarization  models. Our model \combined{}, which combines graph-based, topic-based and thumbnail-based summarization, outperforms all other summarizers.
We further find that the genres RnB and Country are highly overrepresented in the lyrics sample with respect to the full \wasabicorpus{}, indicating that songs belonging to these genres are more likely to contain a chorus.
Finally, Figure~\ref{fig:summary} shows an example summary of four lines length obtained with our proposed \combined{} method. It is toggled in the \wasabinavigator{} by clicking on the green square on top of the song text.

The four-line summaries of 50k English used in our experiments are freely available within the \wasabicorpus{}; the Python code of the applied summarization methods is also available\footnote{\url{https://github.com/TuringTrain/lyrics\_thumbnailing}}.
\section{Explicit Language in Lyrics}
\label{sec:explictness}
On audio recordings, the Parental Advisory Label is placed in recognition of profanity and to warn parents of material potentially unsuitable for children. Nowadays, such labelling is carried out mainly manually on voluntary basis, with the drawbacks of being time consuming and therefore costly, error prone and partly a subjective task. In \cite{fell:explicitness} we have tackled the task of automated explicit lyrics detection, based on the songs carrying such a label. We compared automated methods ranging from dictionary-based lookup to state-of-the-art deep neural networks to automatically detect explicit contents in English lyrics. More specifically, the dictionary-based methods rely on a swear word dictionary $D_n$ which is automatically created from example explicit and clean lyrics. Then, we use $D_n$ to predict the class of an unseen song text in one of two ways: (i) the \textit{Dictionary Lookup} simply checks if a song text contains words from $D_n$. (ii) the \textit{Dictionary Regression} uses BOW made from $D_n$ as the feature set of a logistic regression classifier. In the \textit{Tf-idf BOW Regression} the BOW is expanded to the whole vocabulary of a training sample instead of only the explicit terms. Furthermore, the model \textit{TDS Deconvolution} is a deconvolutional neural network \cite{vanni2018textual} that estimates the importance of each word of the input for the classifier decision.
In our experiments, we worked with 179k lyrics that carry gold labels provided by Deezer (17k tagged as explicit) and obtained the results shown in Figure~\ref{tab:perception:results}. We found the very simple \textit{Dictionary Lookup} method to perform on par with much more complex models such as the \textit{BERT Language Model} \cite{bert} as a text classifier. Our analysis revealed that some genres are highly overrepresented among the explicit lyrics. Inspecting the automatically induced explicit words dictionary reflects that genre bias. The dictionary of 32 terms used for the dictionary lookup method consists of around 50\% of terms specific to the Rap genre, such as glock, gat, clip (gun-related), thug, beef, gangsta, pimp, blunt (crime and drugs). Finally, the terms holla, homie, and rapper are obviously no swear words, but highly correlated with explicit content lyrics.

\begin{table}
    \centering
    \begin{tabular}{lccc}
    \textit{Model} & \textit{P} & \textit{R} & \textit{$F_1$} \\
    \hline
    Majority Class & 45.0 & 50.0 & 47.4 \\
    \hline
    Dictionary Lookup     & 78.3  & 76.4  & 77.3 \\ 
    Dictionary Regression   & 76.2  & 81.5  & 78.5\\ 
    Tf-idf BOW Regression & 75.6  & 81.2  & 78.0\\
    TDS Deconvolution                 & 81.2  & 78.2  & 79.6\\
    BERT Language Model & 84.4  & 73.7  & 77.7\\
    \hline
    \end{tabular}
    \caption{Performance comparison of our different models. Precision (\textit{P}), Recall (\textit{R}) and f-score ($F_1$) in \%.}
    \label{tab:perception:results}
\end{table}

Our corpus contains 52k tracks labelled as explicit and 663k clean (not explicit) tracks\footnote{Labels provided by Deezer. Furthermore, 625k songs have a different status such as unknown or censored version.}. We have trained a classifier (77.3\% f-score on test set) on the 438k English lyrics which are labelled and classified the remaining 455k previously untagged English tracks. We provide both the predicted labels in the \wasabicorpus{} and the trained classifier to apply it to unseen text.
\section{Emotional Description}
\label{sec:emotion}
\begin{figure}
    \centering
    \includegraphics[width=\linewidth]{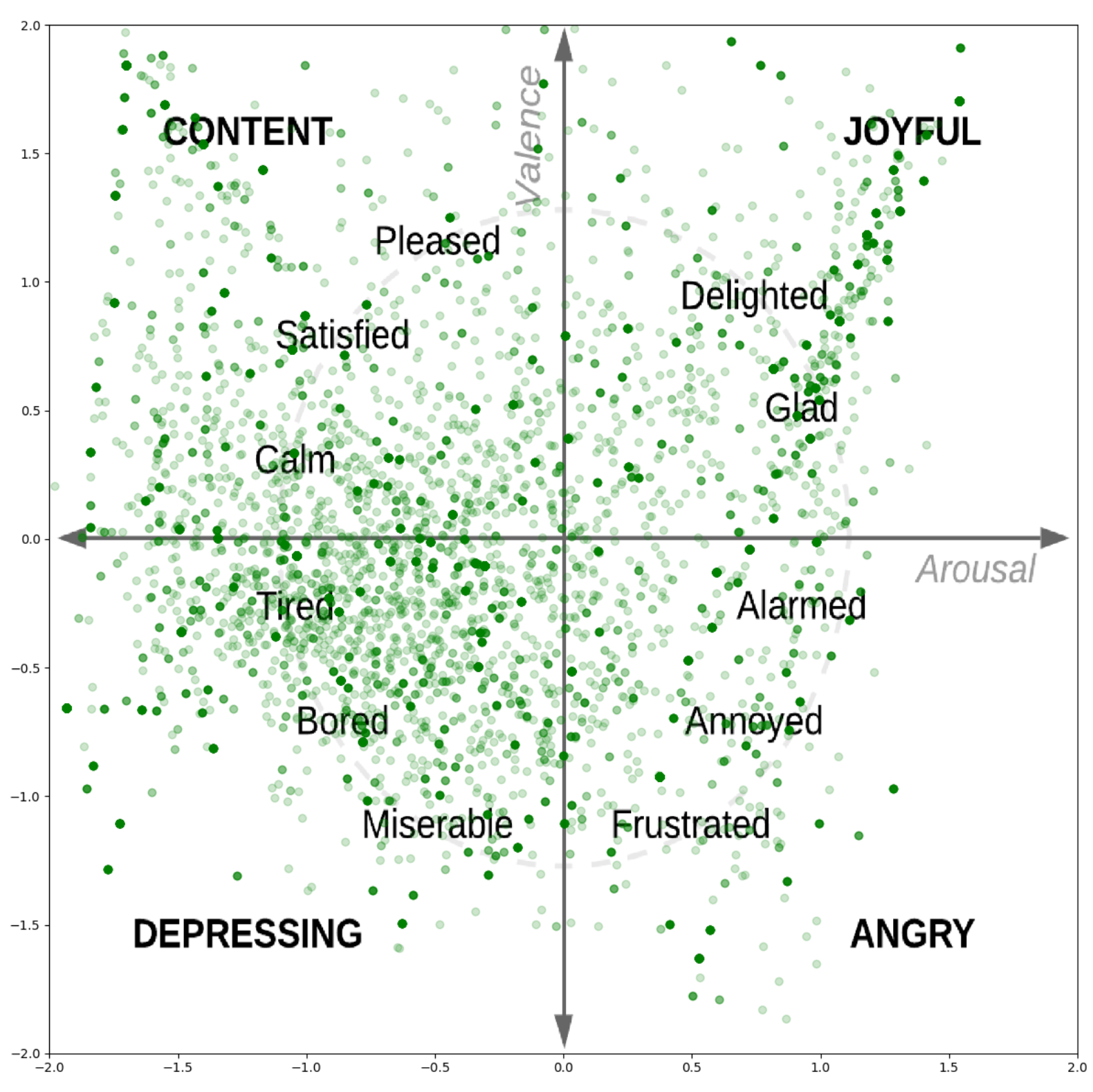}
    \caption{Emotion distribution in the corpus in the valence-arousal plane.}
    \label{fig:emotion_distribution}
\end{figure}

In sentiment analysis the task is to predict if a text has a positive or a negative emotional valence. In the recent years, a transition from detecting sentiment (positive vs. negative valence) to more complex formulations of emotion detection (e.g. joy, fear, surprise) \cite{mohammad2018semeval} has become more visible; even tackling the problem of emotion in context \cite{chatterjee2019semeval}. One family of emotion detection approaches is based on the valence-arousal model of emotion \cite{russell1980circumplex}, locating every emotion in a two-dimensional plane based on its valence (positive vs. negative) and arousal (aroused vs. calm).\footnote{Sometimes, a third dimension of dominance is part of the model.} Figure~\ref{fig:emotion_distribution} is an illustration of the valence-arousal model of Russell and shows exemplary where several emotions such as joyful, angry or calm are located in the plane.
Manually labelling texts with multi-dimensional emotion descriptions is an inherently hard task. Therefore, researchers have resorted to distant supervision, obtaining gold labels from social tags from lastfm. These approaches \cite{hu2009lyric,cano_mood_dataset_creation} define a list of social tags that are related to emotion, then project them into the valence-arousal space using an emotion lexicon \cite{warriner2013norms,mohammad2018obtaining}.

Recently, Deezer made valence-arousal annotations for 18,000 English tracks available\footnote{\url{https://github.com/deezer/deezer\_mood\_detection\_dataset}} they have derived by the aforementioned method \cite{deezer_mm_mood_2018}. We aligned the valence-arousal annotations of Deezer to our songs. In Figure~\ref{fig:emotion_distribution} the green dots visualize the emotion distribution of these songs.\footnote{Depiction without scatterplot taken from \cite{valence_arousal_illustration}} Based on their annotations, we train an emotion regression model using BERT, with an evaluated 0.44/0.43 Pearson correlation/Spearman correlation for valence and 0.33/0.31 for arousal on the test set.

We integrated Deezer's labels into our corpus and also provide the  valence-arousal predictions for the 1.73M tracks with lyrics. We also provide the last.fm social tags (276k) and emotion tags (87k entries) to facilitate researchers to build variants of emotion recognition models.
\section{Topic Modelling}
\label{sec:topics}
We built a topic model on the lyrics of our corpus using Latent Dirichlet Allocation (LDA) \cite{lda}. We determined the hyperparameters $\alpha$, $\eta$ and the topic count such that the coherence was maximized on a subset of 200k lyrics. We then trained a topic model of 60 topics on the unique English lyrics (1.05M).

\begin{figure}
    \centering
    \begin{minipage}{.5\linewidth}
      \centering
      \includegraphics[width=\linewidth]{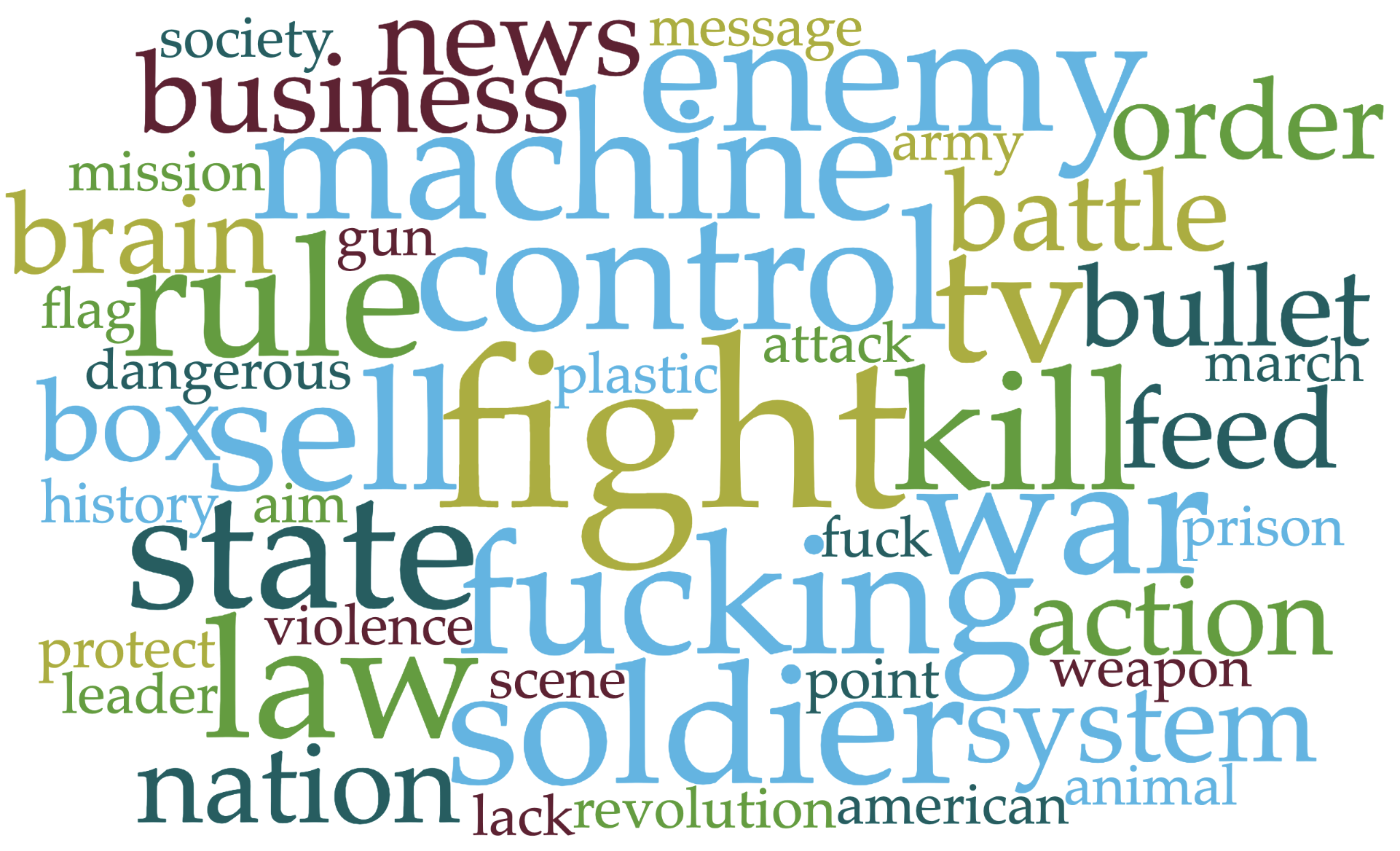}
      \captionof{figure}{Topic War}
      \label{fig:topic_death}
      \vspace{2ex}
    \end{minipage}%
    \begin{minipage}{.5\linewidth}
      \centering
      \includegraphics[width=\linewidth]{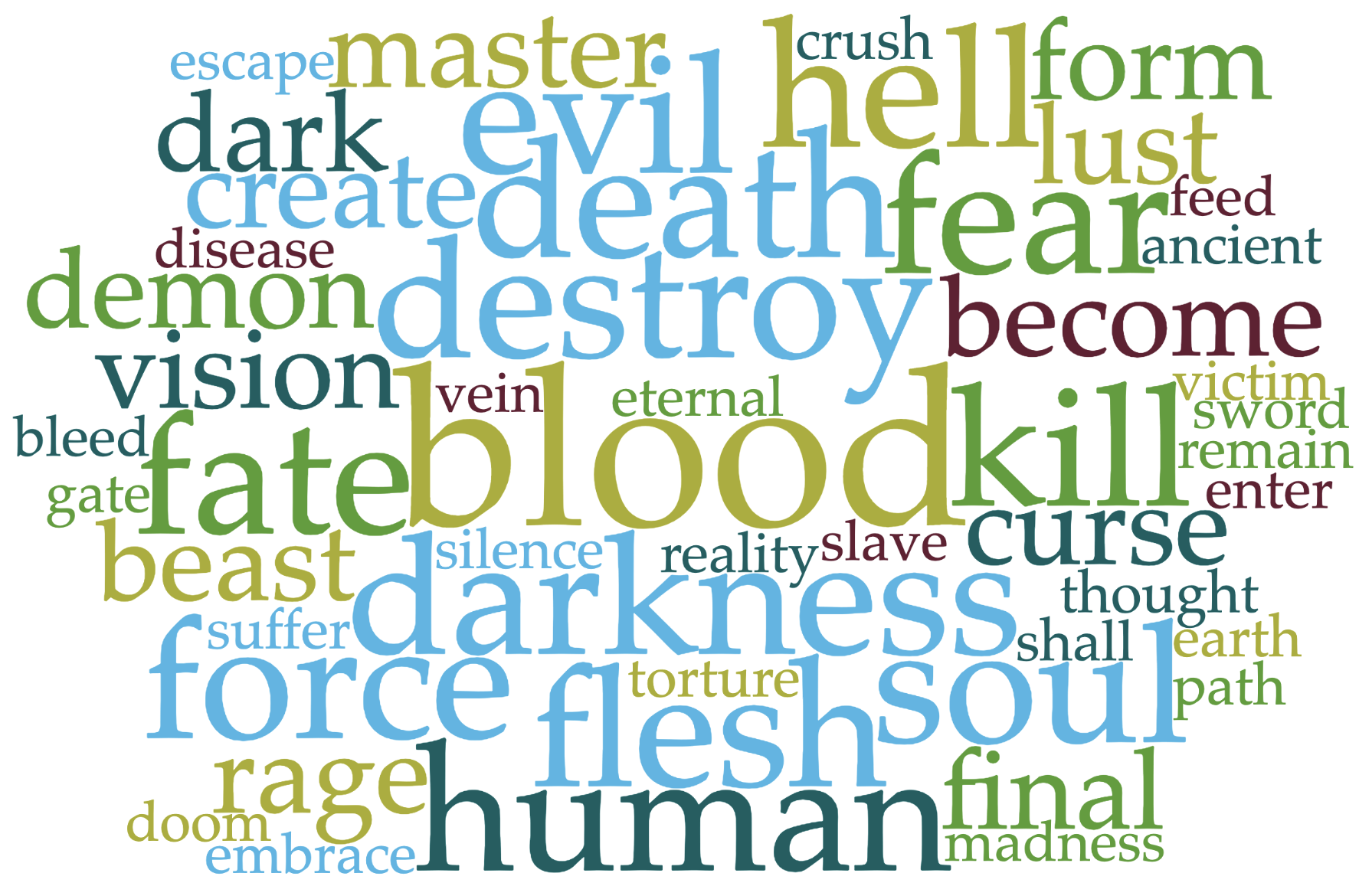}
      \captionof{figure}{Topic Death}
      \label{fig:topic_war}
      \vspace{2ex}
    \end{minipage}
    \begin{minipage}{.5\linewidth}
      \centering
      \includegraphics[width=\linewidth]{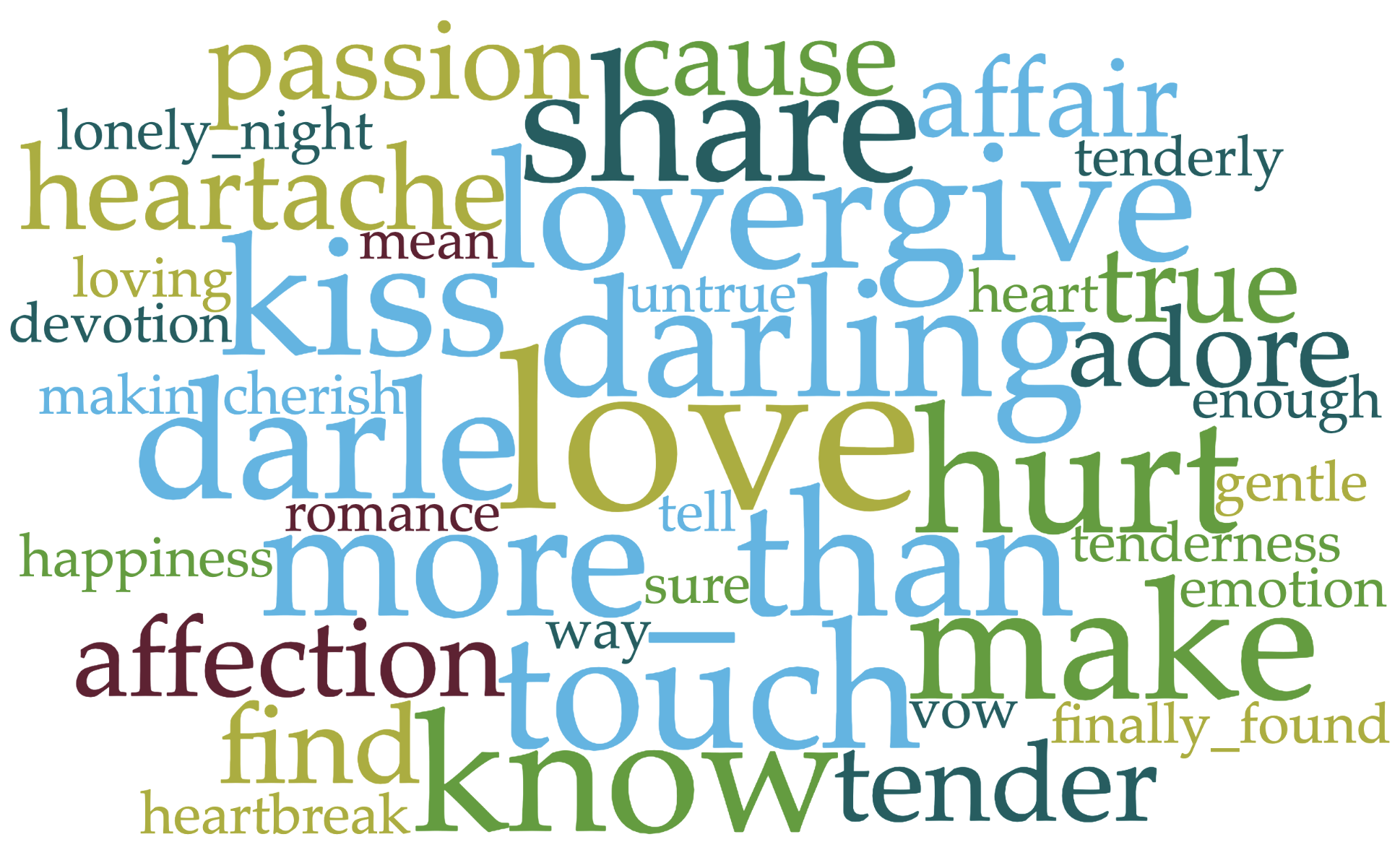}
      \captionof{figure}{Topic Love}
      \label{fig:topic_love}
      \vspace{2ex}
    \end{minipage}%
    \begin{minipage}{.5\linewidth}
      \centering
      \includegraphics[width=\linewidth]{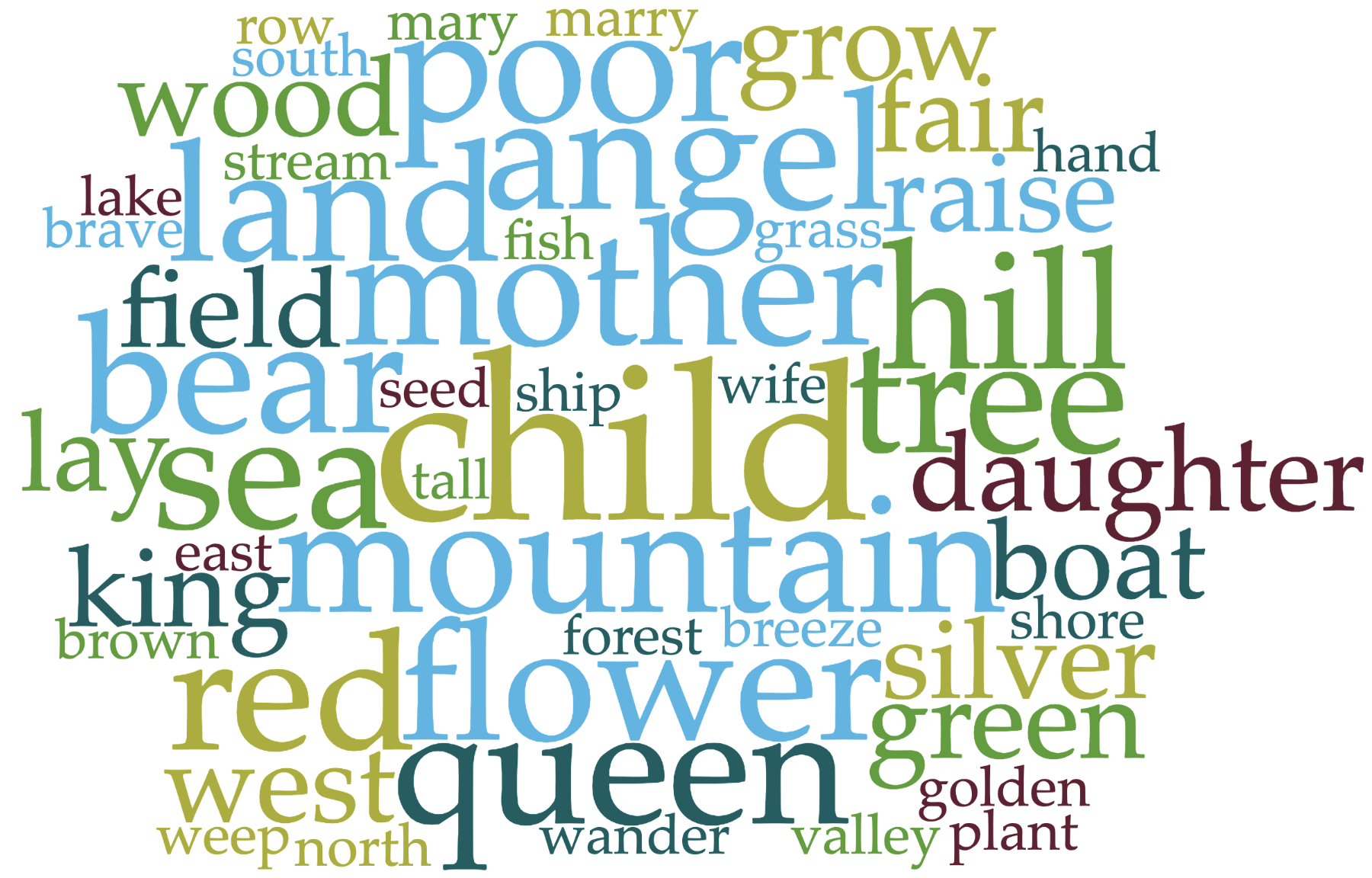}
      \captionof{figure}{Topic Family}
      \label{fig:topic_home}
      \vspace{2ex}
    \end{minipage}
    \begin{minipage}{.5\linewidth}
      \centering
      \includegraphics[width=\linewidth]{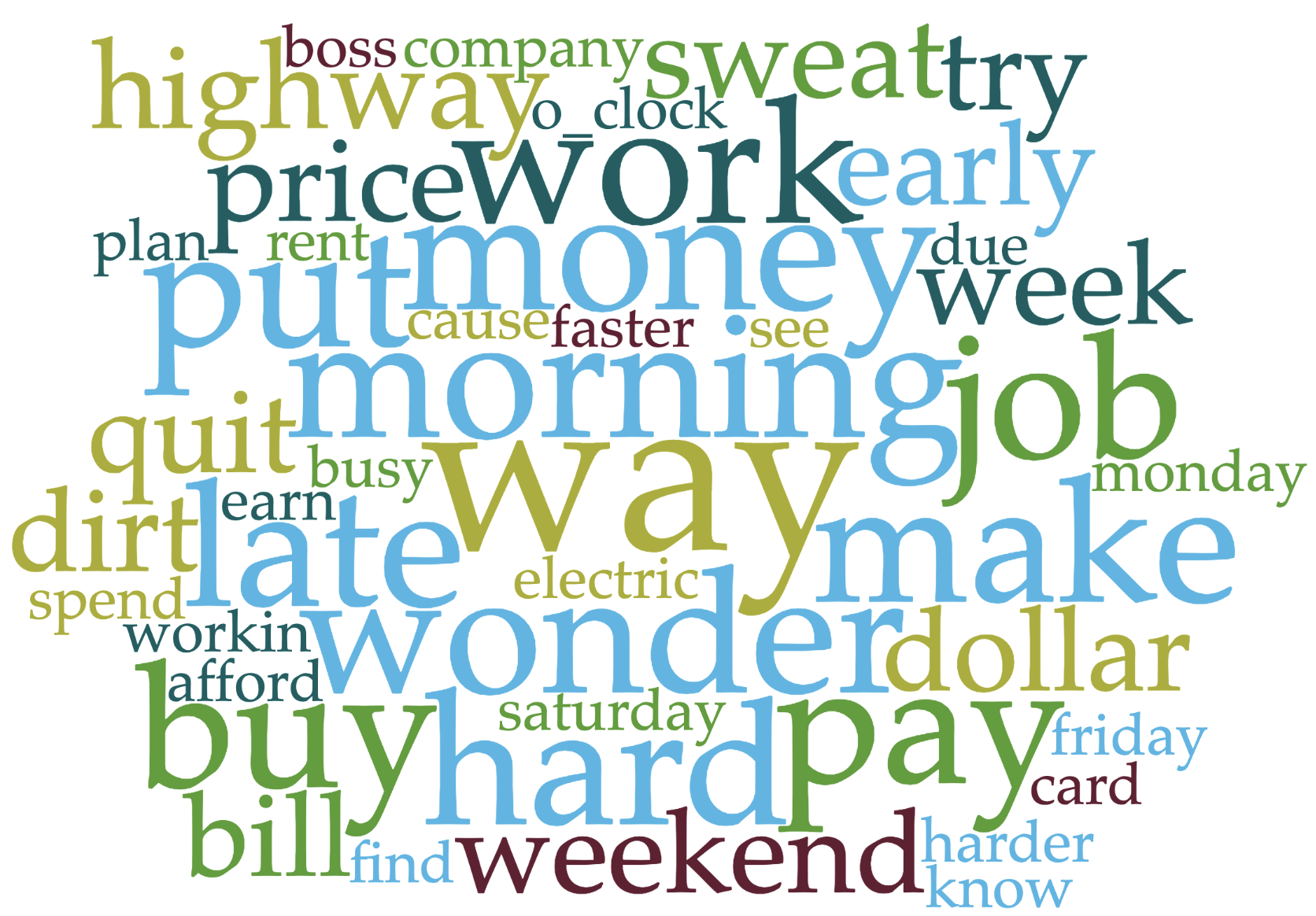}
      \captionof{figure}{Topic Money}
      \label{fig:topic_money}
    \end{minipage}%
    \begin{minipage}{.5\linewidth}
      \centering
      \includegraphics[width=\linewidth]{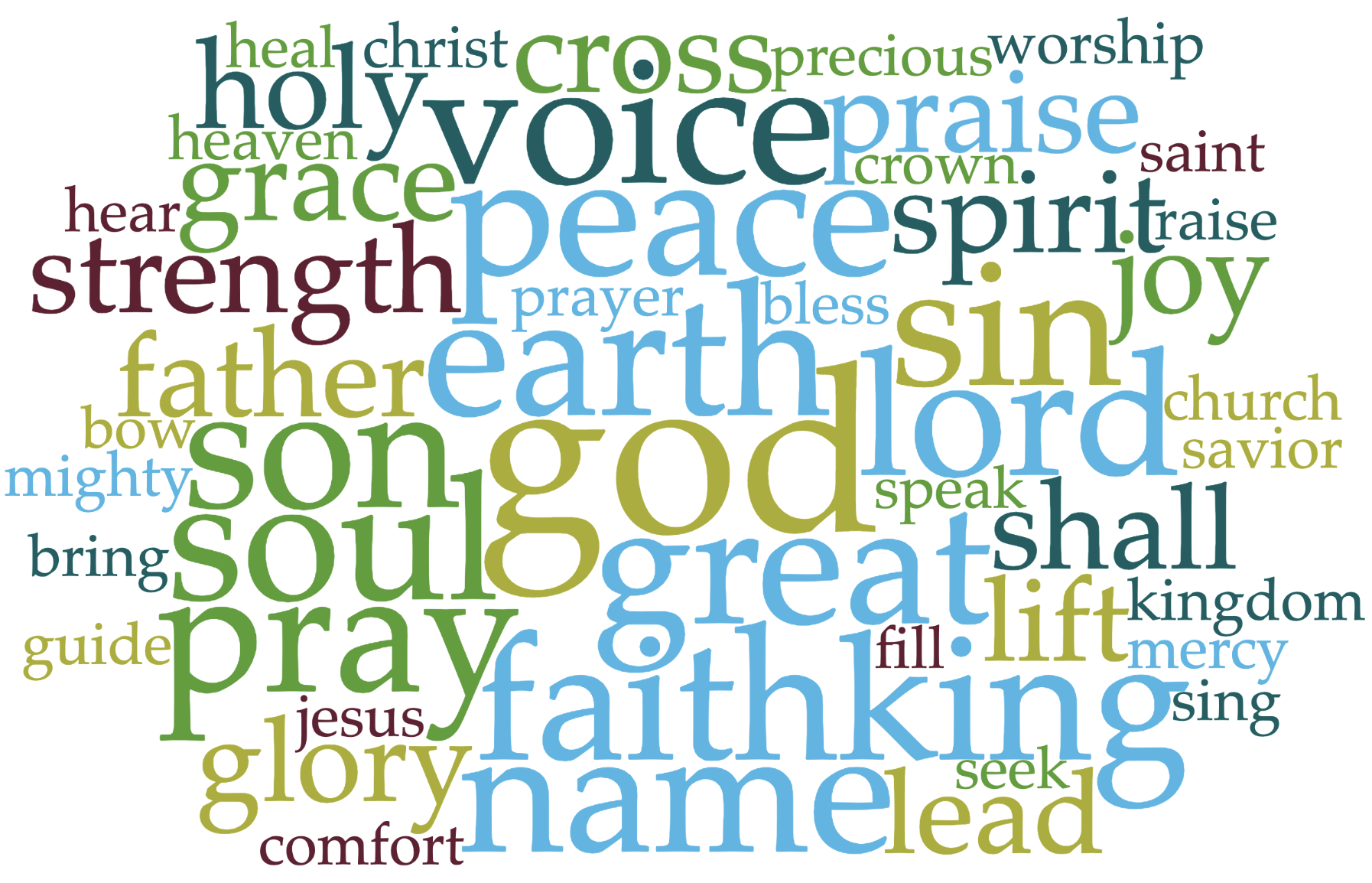}
      \captionof{figure}{Topic Religion}
      \label{fig:topic_religion}
    \end{minipage}
\end{figure}

We have manually labelled a number of more recognizable topics. Figures~\ref{fig:topic_war}-\ref{fig:topic_religion} illustrate these topics with word clouds\footnote{made with \url{https://www.wortwolken.com/}} of the most characteristic words per topic.
For instance, the topic Money contains words of both the field of earning money (job, work, boss, sweat) as well as spending it (pay, buy). The topic Family is both about the people of the family (mother, daughter, wife) and the land (sea, valley, tree).

We provide the topic distribution of our LDA topic model for each song and make available the trained topic model to enable its application to unseen lyrics.
\section{Diachronic Corpus Analysis}
\label{sec:time_analysis}
We examine the changes in the annotations over the course of time by grouping the corpus into decades of songs according to the distribution shown in Figure~\ref{fig:years}.

\begin{figure}
\centering
    \subfloat[Evolution of topic importance
\label{fig:topic_evolution}]{%
      \includegraphics[width=\linewidth]{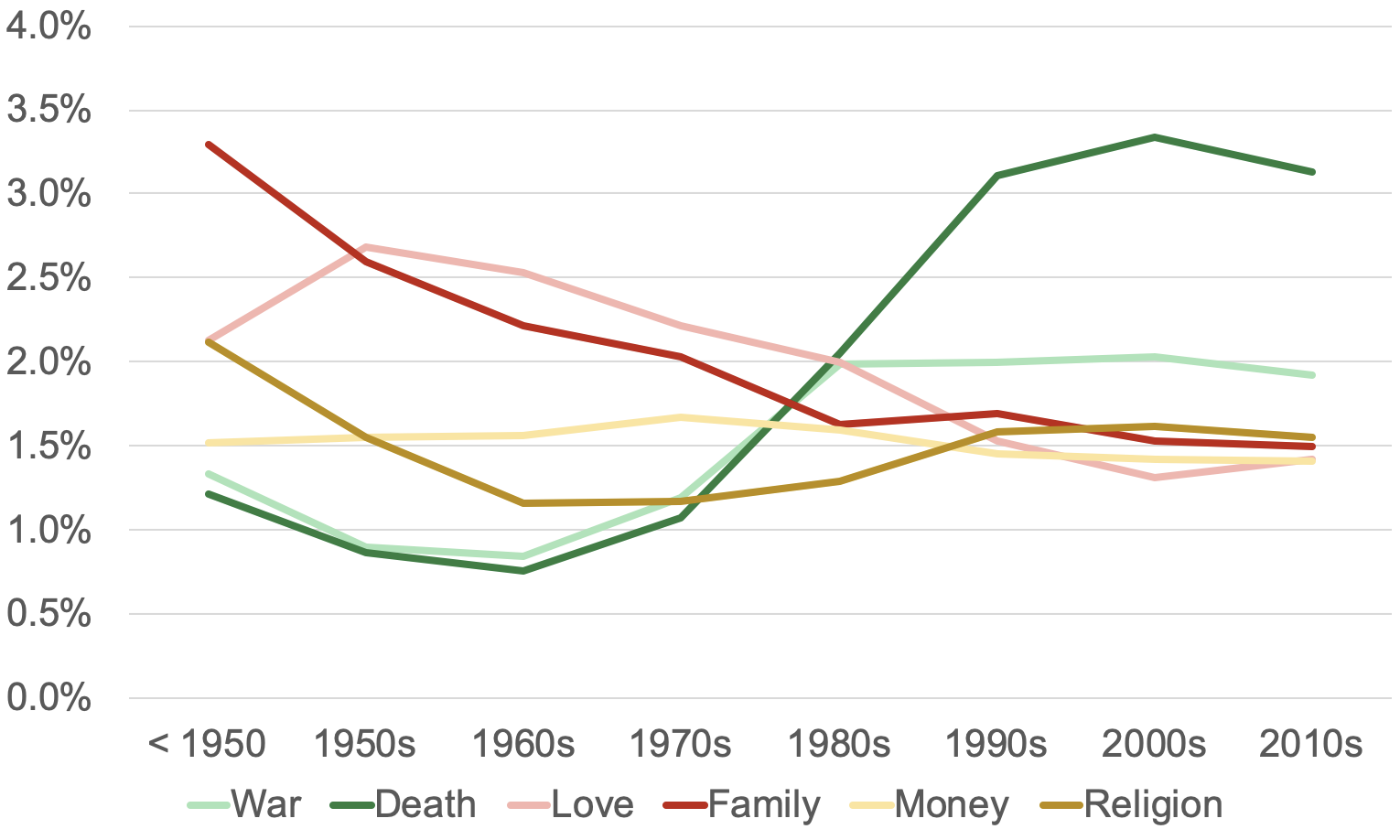}%
      }\par\vspace{4ex} 
    \subfloat[Evolution of explicit content lyrics\label{fig:explicit_evolution}]{%
      \includegraphics[width=\linewidth]{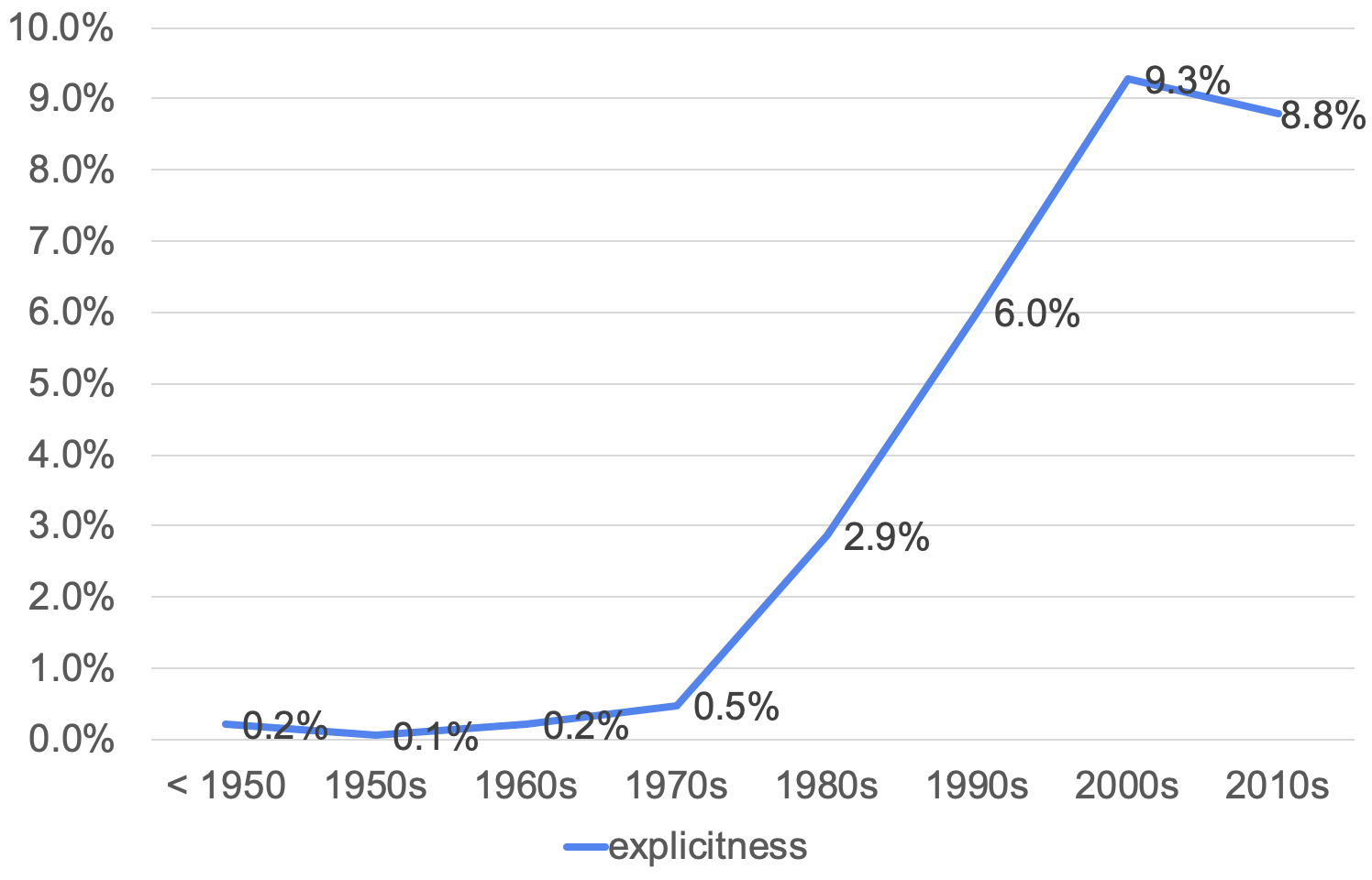}%
      }\par\vspace{4ex}      
    \subfloat[Evolution of emotion\label{fig:emotion_evolution}]{%
      \includegraphics[width=\linewidth]{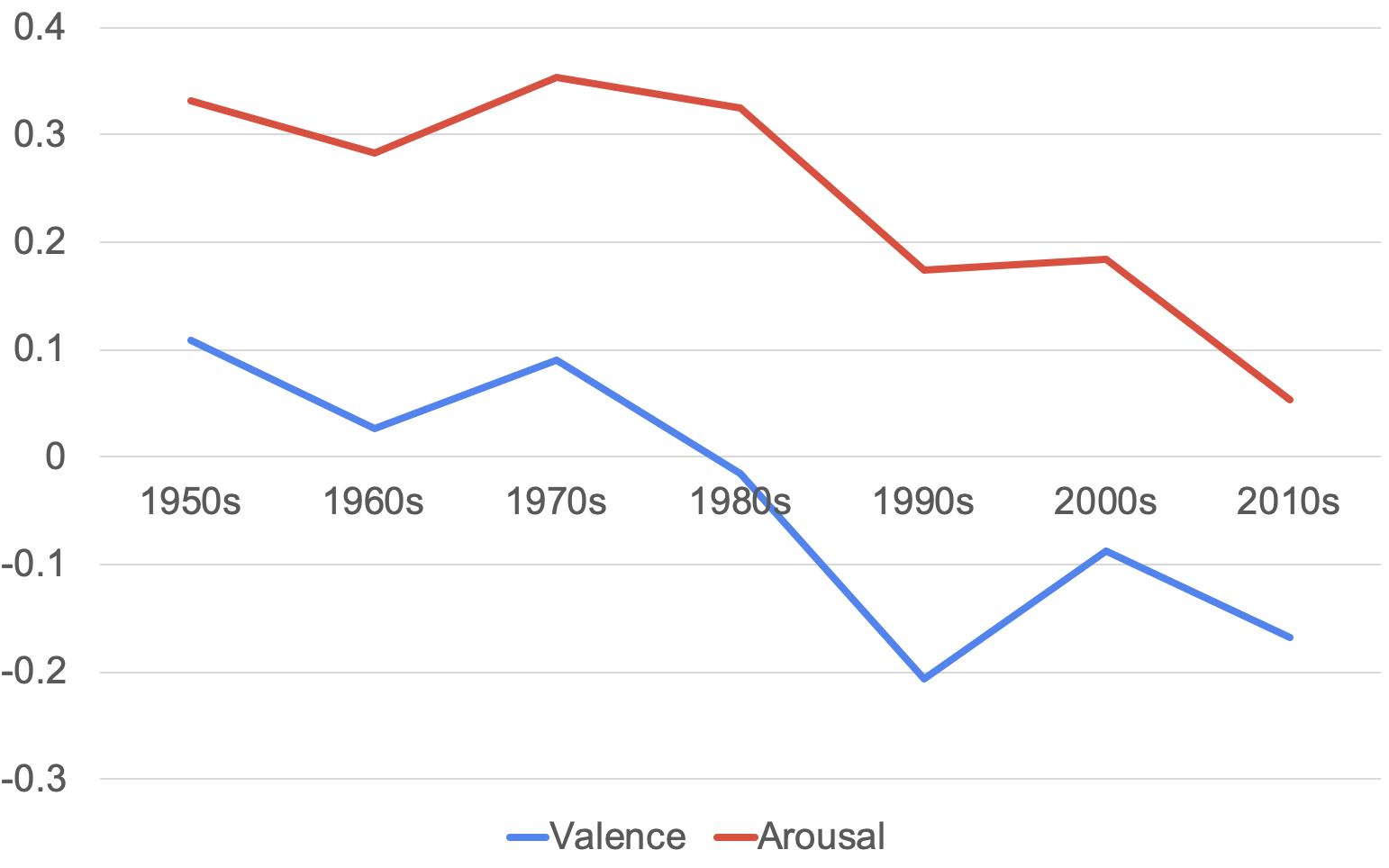}%
      }
\caption{Evolution of different annotations during the decades}
\label{fig:evolution}
\end{figure}

\paragraph{Changes in Topics}
\label{sec:topic_evolution}
The importance of certain topics has changed over the decades, as depicted in Figure~\ref{fig:topic_evolution}. Some topics have become more important, others have declined, or stayed relatively the same. We define the importance of a topic for a decade of songs as follows: first, the LDA topic model trained on the full corpus gives the probability of the topic for each song separately. We then average these song-wise probabilities over all songs of the decade.
For each of the cases of growing, diminishing and constant importance, we display two topics. The topics War and Death have appreciated in importance over time. This is partially caused by the rise of Heavy Metal in the beginning of the 1970s, as the vocabulary of the Death topic is very typical for the genre (see for instance the \qu{Metal top 100 words} in \cite{fell2014lyrics}). We measure a decline in the importance of the topics Love and Family. The topics Money and Religion seem to be evergreens as their importance stayed rather constant over time.
\paragraph{Changes in Explicitness}
We find that newer songs are more likely being tagged as having explicit content lyrics. Figure~\ref{fig:explicit_evolution} shows our estimates of explicitness per decade, the ratio of songs in the decade tagged as explicit to all songs of the decade. Note that the Parental Advisory Label was first distributed in 1985 and many older songs may not have been labelled retroactively. The depicted evolution of explicitness may therefore overestimate the \qu{true explicitness} of newer music and underestimate it for music before 1985.
\paragraph{Changes in Emotion}
We estimate the emotion of songs in a decade as the average valence and arousal of songs of that decade. We find songs to decrease both in valence and arousal over time.
This decrease in positivity (valence) is in line with the diminishment of positively connotated topics such as Love and Family and the appreciation of topics with a more negative connotation such as War and Death.

\section{Related Work} \label{sec:related}
This section describes available songs and lyrics databases, and summarizes existing work on lyrics processing.

\paragraph{Songs and Lyrics Databases.}
The Million Song Dataset (MSD) project\footnote{\url{http://millionsongdataset.com}} \cite{Bertin-Mahieux2011} is a collection of audio features and metadata for a million contemporary popular music tracks. Such dataset shares some similarities with WASABI with respect to metadata extracted from Web resources (as artist names, tags, years) and audio features, even if at a smaller scale. Given that it mainly focuses on audio data, a complementary dataset providing lyrics of the Million Song dataset was released, called musiXmatch dataset\footnote{\url{http://millionsongdataset.com/musixmatch/}}. It consists in a collection of song lyrics in bag-of-words (plus stemmed words), associated with MSD tracks. However, no other processing of the lyrics is done, as is the case in our work.

MusicWeb and its successor MusicLynx \cite{allik2018www} link music artists within a Web-based application for discovering connections between them and provides a browsing experience using extra-musical relations. The project shares some ideas with WASABI, but works on the artist level, and does not perform analyses on the audio and lyrics content itself. It reuses, for example, MIR metadata from AcousticBrainz.

The WASABI project has been built on a broader scope than these projects and mixes a wider set of metadata, including ones from audio and natural language processing of lyrics. In addition, as presented in this paper, it comes with a large set of Web Audio enhanced applications (multitrack player, online virtual instruments and effect, on-demand audio processing, audio player based on extracted, synchronized chords, etc.)

Companies such  as  Spotify,  GraceNote, Pandora,  or Apple  Music  have sophisticated private knowledge bases of songs and lyrics to feed their search and recommendation algorithms, but such data are not available (and mainly rely on audio features).

\paragraph{Lyrics Segmentation.} Only a few papers in the literature have focused on the automated detection of the structure of lyrics. \cite{watanabe2016modeling} propose the task to automatically identify segment boundaries in lyrics and train a logistic regression model for the task with the repeated pattern and textual features.
\cite{Mahedero:2005:NLP:1101149.1101255} report experiments on the use of standard NLP tools for the analysis of music lyrics. Among the tasks they address, for structure extraction they focus on a small sample of lyrics having a clearly recognizable structure (which is not always the case) divided into segments. 
More recently, \cite{6735570} describe a semantics-driven approach to the automatic segmentation of song lyrics, and mainly focus on pop/rock music. Their goal is not to label a set of lines in a given way (e.g. verse, chorus), but rather identifying recurrent as well as non-recurrent groups of lines. They propose a rule-based method to estimate such structure labels of segmented lyrics.

\paragraph{Explicit Content Detection.}
\cite{bergelid2018classification} consider a dataset of English lyrics to which they apply classical machine learning algorithms.
The explicit labels are obtained from Soundtrack Your Brand~\footnote{\url{https://www.soundtrackyourbrand.com}}.
They also experiment with adding lyrics metadata to the feature set, such as the artist name, the release year, the music energy level, and the valence/positiveness of a song. 
\cite{chin2018explicit} apply explicit lyrics detection to Korean song texts. They also use tf-idf weighted BOW as lyrics representation and aggregate multiple decision trees via boosting and bagging to classify the lyrics for explicit content.
More recently, \cite{kim2019hybrid} proposed a neural network method to create explicit words dictionaries automatically by weighting a vocabulary according to all words' frequencies in the explicit class vs. the clean class, accordingly.
They work with a corpus of Korean lyrics.

\paragraph{Emotion Recognition}
Recently, \cite{deezer_mm_mood_2018} address the task of multimodal music mood prediction based on the audio signal and the lyrics of a track. They propose a new model based on deep learning outperforming traditional feature engineering based approaches. Performances are evaluated on their published dataset with associated valence and arousal values which we introduced in Section~\ref{sec:emotion}

\cite{Xia:2008:SVS:1557690.1557725} model song texts in a low-dimensional vector space as bags of concepts, the \qu{emotional units}; those are combinations of
emotions, modifiers and negations.
\cite{5363083} leverage the music's emotion annotations from Allmusic which they map to a lower dimensional psychological model of emotion. They train a lyrics emotion classifier and show by qualitative interpretation of an ablated model (decision tree) that the deciding features leading to the classes are intuitively plausible.
\cite{Hu2009LyricbasedSE} aim to detect emotions in song texts based on Russell's model of mood; rendering emotions
continuously in the two dimensions of arousal and valence (positive/negative). They analyze each sentence as bag of \qu{emotional units}; they reweight sentences' emotions by both adverbial modifiers and tense and even consider
progressing and adversarial valence in consecutive sentences. Additionally, singing speed is taken into account.
With the fully weighted sentences, they perform clustering in the 2D plane of valence and arousal. Although
the method is unsupervised at runtime, there are many parameters tuned manually by the authors in this work.

\cite{mihalcea-strapparava-2012-lyrics} render emotion detection as a multi-label classification problem, songs express intensities of six different basic emotions: anger, disgust, fear, joy, sadness, surprise. Their corpus (100 song
texts) has time-aligned lyrics with information on musical key and note progression. Using Mechanical Turk they each line of song text is annotated with the six emotions. For emotion classification, they use bags of words and concepts, as musical features key and notes. Their classification results using both modalities, textual and audio features, are significantly improved compared to a single modality.

\paragraph{Topic Modelling}
Among the works addressing this task for song lyrics, \cite{Mahedero:2005:NLP:1101149.1101255} define five ad hoc topics (Love, Violent, Antiwar, Christian, Drugs) into which they classify their corpus of 500 song texts using supervision. Related, \cite{fellthesis} also use supervision to find bags of genre-specific n-grams. Employing the view from the
literature that BOWs define topics, the genre-specific terms can be seen as mixtures of genre-specific topics.

\cite{1394328} apply the unsupervised topic model Probabilistic LSA to their ca. 40k song texts. They learn latent topics for both the lyrics corpus as well as a NYT newspaper corpus (for control) and show that the domain-specific topics slightly improve the performance in their MIR task. While their MIR task performs highly better when using acoustic features, they discover that
both methods err differently. 
\cite{Kleedorfer2008OhOO} apply Non-negative Matrix Factorization (NMF) to ca. 60k song texts and cluster them into 60 topics. They show the so discovered topics to be intrinsically meaningful.

\cite{sterckx2014topic} have worked on topic modelling of a large-scale lyrics corpus of 1M songs. They build models using Latent Dirichlet allocation with topic counts between 60 and 240 and show that the 60 topics model gives a good trade-off between topic coverage and topic redundancy.
Since popular topic models such as LDA represent topics as weighted bags of words, these topics are not immediately interpretable. This gives rise to the need of an automatic labelling of topics with smaller labels. A recent approach \cite{auto_topic_labelling} relates the topical BOWs with titles of Wikipedia articles in a two step procedure: first, candidates are generated, then ranked. 
\section{Conclusion} \label{sec:conclusion}
In this paper we have described the WASABI dataset of songs, focusing in particular on the lyrics annotations resulting from the applications of the methods we proposed to extract relevant information from the lyrics. So far, lyrics annotations concern their
structure segmentation, their topic, the explicitness of the lyrics content, the summary of a song and the emotions conveyed. Some of those annotation layers are provided for all the 1.73M songs included in the WASABI corpus, while some others apply to subsets of the corpus, due to various constraints described in the paper. 
Table~\ref{tab:dataset_overview} summarizes the most relevant annotations in our corpus.

\begin{table}[ht!]
    \centering
    \begin{tabular}{lll}
    \textit{Annotation} & \textit{Labels} & \textit{Description}\\
    \hline
    Lyrics & 1.73M & segments of lines of text\\
    Languages & 1.73M & 36 different ones\\
    Genre & 1.06M & 528 different ones\\
    Last FM id & 326k & UID \\
    Structure & 1.73M & SSM $\in \mathbb{R}^{n \times n}$ (n: length)\\
    $\mathbb{S}$ocial tags & 276k & $\mathbb{S}=\text{\{rock, joyful, 90s, ...\}}$\\
    $\mathbb{E}$motion tags & 87k & $\mathbb{E} \subset \mathbb{S}=\text{\{joyful, tragic, ...\}}$\\
    Explicitness & 715k & True (52k), False (663k)\\
    Explicitness \textsuperscript{\ding{168}} & 455k & True (85k), False (370k)\\
    Summary\textsuperscript{\ding{168}} & 50k & four lines of song text\\
    Emotion & 16k & (valence, arousal) $\in \mathbb{R}^{2}$\\
    Emotion\textsuperscript{\ding{168}} & 1.73M & (valence, arousal) $\in \mathbb{R}^{2}$\\
    Topics\textsuperscript{\ding{168}} & 1.05M & Prob. distrib. $\in \mathbb{R}^{60}$\\
    \hline
    Total tracks & 2.10M & diverse metadata\\
    \hline
    \end{tabular}
    \caption{Most relevant song-wise annotations in the \wasabicorpus{}. Annotations with \ding{168} are predictions of our models.}
    \label{tab:dataset_overview}
\end{table}

As the creation of the resource is still ongoing, we plan to integrate an improved emotional description in future work. In \cite{influence_networks} the authors have studied how song writers influence each other. We aim to learn a model that detects the border between heavy influence and plagiarism.

\section*{Acknowledgement}
This work is partly funded by the French  Research  National Agency (ANR) under the WASABI project (contract ANR-16-CE23-0017-01) and by the EU Horizon 2020 research and innovation programme under the Marie Sklodowska-Curie grant agreement No. 690974 (MIREL).

\section{References}
\bibliographystyle{lrec}
\bibliography{ref}

\end{document}